\documentclass[conference]{IEEEtran}
  \usepackage{mathtools,bm,amsfonts,amssymb,algorithm,algorithmicx,caption,algpseudocode,mathrsfs,multirow,multicol,subcaption,booktabs,threeparttable,balance,hyperref,breqn,dsfont,enumitem,cite}

\newcommand{\sysname}{\textsc{MLAS} }
\newcommand{\sysnamenospace}{\textsc{MLAS}}
\newcommand{\anet}{\texttt{AttNet} }
\newcommand{\anetns}{\texttt{AttNet}}
\newcommand{\snet}{\texttt{SeqNet} }
\newcommand{\snetns}{\texttt{SeqNet}}
\newcommand{\fnet}{\texttt{FusionNet} }
\newcommand{\fnetns}{\texttt{FusionNet}}
\newcommand{\mnet}{\texttt{MetricNet} }
\newcommand{\mnetns}{\texttt{MetricNet}}

\newcommand{\ie}{\emph{i.e.}}
\newcommand{\eg}{\emph{e.g.}}
\newcommand{\etc}{\emph{etc}}
\newcommand{\figwidth}{0.24}

\newtheorem{definition}{Definition} %
\DeclareMathOperator*{\minimize}{minimize}

\makeatletter
\newcommand*\bigcdot{\mathpalette\bigcdot@{.8}}
\newcommand*\bigcdot@[2]{\mathbin{\vcenter{\hbox{\scalebox{#2}{$\m@th#1\bullet$}}}}}
\makeatother
\newcommand*{\thead}[1]{\multicolumn{1}{c}{#1}}

\usepackage{varwidth}
\DeclareCaptionFormat{centercaption}{%
  \begin{varwidth}{\linewidth}%
    \centering
    #1#2#3%
  \end{varwidth}%
}
  
\renewcommand{\vec}[1]{\bm{#1}}
\algnewcommand\algorithmicinput{\textbf{INPUT:}}
\algnewcommand\INPUT{\item[\algorithmicinput]}
\algnewcommand\algorithmicoutput{\textbf{OUTPUT:}}
\algnewcommand\OUTPUT{\item[\algorithmicoutput]}
\algnewcommand\algorithmicforeach{\textbf{for each}}
\algdef{S}[FOR]{ForEach}[1]{\algorithmicforeach\ #1\ \algorithmicdo}
   \newif\ifoverleaf
  \overleaftrue
  \newif\iflocalbuild
  \localbuildfalse
  \newif\ifredcomment
  \redcommentfalse

    \IEEEpubid{978-1-7281-6251-5/20/\$31.00~\copyright2020~IEEE } 
    
\begin{document}
    \title{MLAS: \underline{M}etric \underline{L}earning on \underline{A}ttributed \underline{S}equences}
    \author{\IEEEauthorblockN{Zhongfang Zhuang, Xiangnan Kong, Elke Rundensteiner}
              \IEEEauthorblockA{Worcester Polytechnic Institute \\ Worcester, MA 01609 \\ \{zzhuang, xkong, rundenst\}@wpi.edu}
              \and 
              \IEEEauthorblockN{Jihane Zouaoui, Aditya Arora}
              \IEEEauthorblockA{Amadeus \\ Sophia-Antipolis, France \\ \{jihane.zouaoui, aditya.arora\}@amadeus.com}
              }
  \maketitle

\begin{abstract}
Distance metric learning has attracted much attention in recent years, where the goal is to learn a distance metric based on user feedback. 
Conventional approaches to metric learning mainly focus on learning the Mahalanobis distance metric on data attributes. 
Recent research on metric learning has been extended to sequential data, where we only have structural information in the sequences, but no attribute is available. 
However, real-world applications often involve \textit{attributed sequence} data (e.g., clickstreams), where each instance consists of not only \textit{a set of attributes} (e.g., user session context) but also \textit{a sequence of categorical items} (e.g., user actions). 
In this paper, we study the problem of metric learning on \textit{attributed sequences}. %
Unlike previous work on metric learning, we now need to go beyond the Mahalanobis distance metric in the attribute feature space while also incorporating the structural information in sequences. 
We propose a deep learning framework, called \sysname (\underline{M}etric \underline{L}earning on \underline{A}ttributed \underline{S}equences), to learn a distance metric that effectively measures dissimilarities between attributed sequences. 
Empirical results on real-world datasets demonstrate that the proposed \sysname framework significantly improves the performance of metric learning compared to state-of-the-art methods on attributed sequences. 
\end{abstract}     

\section{Introduction}
\label{intro}
Distance metric learning, where the goal is to learn a distance metric based on a set of \textit{similar/dissimilar} pairs of instances, has attracted significant attention in recent years~\cite{wang2015survey, wang2011integrating, yeung2007kernel, koestinger2012large, cvpr-face-verify, mueller2016siamese}. Many real-world applications from web log analysis to user profiling could significantly benefit from distance metric learning. 

Conventional approaches to distance metric learning~\cite{xing2003distance, davis2007information, koestinger2012large, mignon2012pcca} focus on learning a Mahalanobis distance metric, which is equivalent to learning a \textit{linear transformation} on data attributes. 
Recent research has extended distance metric learning to nonlinear settings using deep neural networks~\cite{NIPS2012_4840, cvpr-face-verify}, where a nonlinear mapping function is first learned to project the instances into a new space, and then the final metric becomes the Euclidean distance metric in that space. With the flexibility and powerfulness demonstrated in various applications, distance metric learning using neural networks has been the method of choice for learning such nonlinear mappings~\cite{wang2012parametric, chatpatanasiri2010new, NIPS2012_4840, cvpr-face-verify, mueller2016siamese}. 
\begin{figure}[t]
    \centering
       \includegraphics[width=0.95\linewidth]{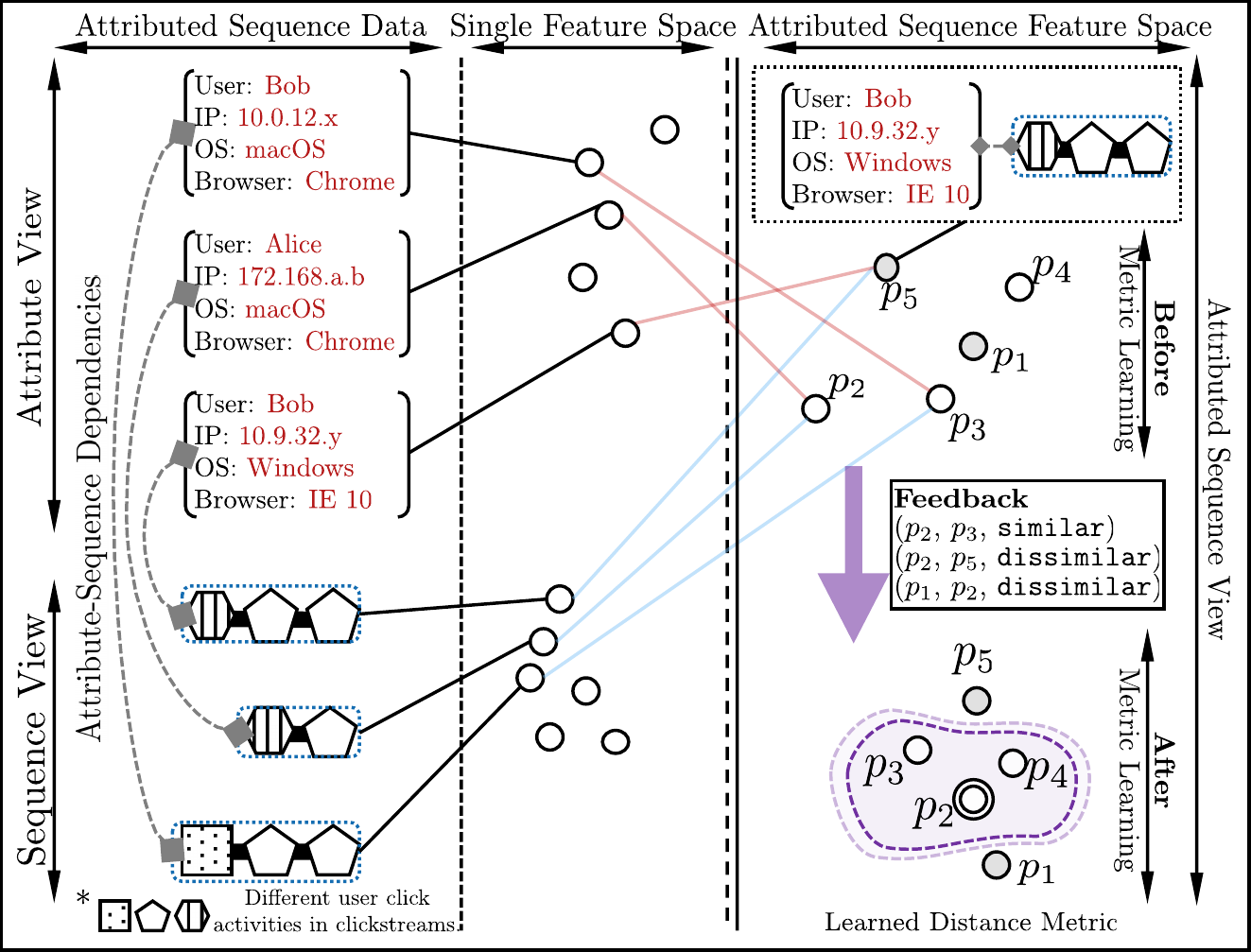}
    \caption{Metric learning on attributed sequences. }
    \label{fig-feedback}
    \vspace{-6mm}
\end{figure}
 
While some recent research on metric learning has begun to explore more complex data, such as sequences~\cite{mueller2016siamese}, web applications often involve the even more complex \textit{attributed sequence}, where each instance consists of not only a sequence of categorical items but also a set of attributes of numerical or categorical values. 
For example, each user browsing session in a bot detection system can be viewed as an attributed sequence, with the series of click activities as the sequence and static context (\eg, browser name, operating system version) as the attributes. 
Furthermore, the attributes and the sequences from the many applications are not independent of each other. For example, in a web search, each user session is composed of a session profile modeled by a set of attributes (\eg, \textit{type of device}, \textit{operating system}, \etc) as well as a sequence of search \textit{keywords}. One \textit{keyword} may depend on previous search terms (\eg, ``\textit{temperature}'' following ``\textit{snow storm}'') and the keywords searched may also depend on the attribute device type (\eg, ``\textit{nearest restaurant}'' on ``\textit{cellphone}''). 

In this paper, we study this new problem of \textit{distance metric learning on attributed sequences}, where the distance between similar attributed sequences is minimized, while dissimilar ones become well-separated with a margin. This problem is core to many applications from fraud detection to user behavior analysis for targeted advertising. This problem is challenging since explicit class labels for attributed sequences are often \textit{not} available. Instead, only a few pairs of attributed sequences may be known to be similar or dissimilar. 
This problem differs from previous works on metric learning~\cite{xing2003distance, wang2011integrating, yeung2007kernel} because we need to go beyond linear transformations in the attribute feature space while incorporating not only the structural information in sequences but also the dependencies between attributes and sequences. 
In this work, we propose a deep learning framework, called \sysname (\underline{M}etric \underline{L}earning on \underline{A}ttributed \underline{S}equences), targeting at learning a distance metric that can effectively exploit the similarity and dissimilarity between each pair of attributed sequences. Our paper offers the following contributions: 
\begin{itemize}
    \item We formulate and study the problem of distance metric learning on attributed sequences. 
    \item We design three main sub-networks to learn the information from the attributes, the sequences, and the attribute-sequence dependencies.
    \item We design three distinct integrated deep learning architectures with three main sub-networks. 
    \item We design an experimental strategy to evaluate and compare the distance metric learned by our proposed \sysname network with various baseline methods. 
\end{itemize}

 We organize the paper as follows. We first define our problem in Section~\ref{section-formulation}. We detail the \sysname network design and the metric learning process in Section~\ref{section-model}. Next, we present the experimental methodology and results in Section~\ref{section-experiments}. We summarize the related work in Section~\ref{section-related}. We conclude our findings and discuss future research directions in Section~\ref{section-conclusion}.
\section{Problem Formulation}
\label{section-formulation}
\subsection{Preliminaries}
\begin{definition}[Sequence]{\rm
Given a set of categorical items $\mathcal{I}=\{e_1, \cdots, e_r\}$, a sequence $S_k = \left( \alpha_k^{(1)}, \cdots, \alpha_k^{(T_k)} \right)$ is an ordered list of $T_k$ items, where the item at $t$-th time step $\alpha_k^{(t)} \in \mathcal{I}, \forall t=1,\cdots, T_k$.} 
\end{definition}
In prior work~\cite{graves2013generating}, one common preprocessing step is to zero-pad each sequence to the longest sequence in the dataset and then to one-hot encode it. Without loss of generality, we denote the length of the longest sequence as $T$. 
For each $k$-th sequence $S_k$ in the dataset, we denote its equivalent one-hot encoded sequence as:
$\mathbf{S}_k = \left( \vec{\alpha}_k^{(1)}, \cdots, \vec{\alpha}_k^{(T)} \right)$, where $\vec{\alpha}_k^{(t)}\in \mathbb{R}^r$ corresponds to a vector represents the item $\alpha_k^{(t)}$ with the $l$-th entry in $\vec{\alpha}_k^{(t)}$ is ``1'' and all other entries are zeros if $\alpha_k^{(t)}=e_l, e_l \in \mathcal{I}$. %

\begin{definition}[Attributed Sequence]{\rm
Given an attribute vector $\mathbf{x}_k\in\mathbb{R}^u$ with $u$ attributes,  
and a sequence $\mathbf{S}_k\in \mathbb{R}^{T \times r}$, an attributed sequence $p_k = (\mathbf{x}_k, \mathbf{S}_k)\in \left(\mathbb{R}^{u}, \mathbb{R}^{T \times r}\right)$ is a pair composed of the attribute vector $\mathbf{x}_k$ and the corresponding sequence $\mathbf{S}_k$. }
\end{definition} 
For example, in a flight booking system, one attributed sequence represents a booking session by the end user. In this case, the attribute vector represents the booking session's profile (\eg, \texttt{IP address}, \texttt{session duration}, \etc) while the sequence consists of the end user's activities on the booking webpage. 
Given a set of $n$ attributed sequences $\mathcal{J} = \{J_1, \cdots, J_n\}$, $\mathbf{S}_k \in \mathbb{R}^{T \times r}$, where $T = \max\{T_1, \cdots, T_n\}$.
Each attribute vector is composed of a number of attributes, where the value of each attribute is in $\mathbb{R}$.  
The dimension of attribute value vectors depends on the number of distinct attribute value combinations. 

We further define the \textit{feedback} as a collection of similar (or dissimilar) attributed sequence pairs. 
\begin{definition}[Feedback]{\rm %
\label{def-as-feedback}
Let $\mathcal{P} = \{ p_1, \cdots, p_n\}$ be a set of $n$ attributed sequences. A feedback is a triplet $(p_i, p_j, \ell_{ij})$ consisting of two attributed sequences $p_i, p_j \in \mathcal{P}$ and a label $\ell_{ij} \in \{0, 1\}$ indicating whether $p_i$ and $p_j$ are similar ($\ell_{ij}=0$) or dissimilar ($\ell_{ij}=1$). We define a similar feedback set $\mathcal{S} = \{(p_i, p_j, \ell_{ij}) | \ell_{ij}=0\}$ and a dissimilar feedback set $\mathcal{D} = \{(p_i, p_j, \ell_{ij}) | \ell_{ij}=1\}$.}
\end{definition}
The feedback could be given by domain experts in real-world applications based on their domain experiences. 

\noindent \textbf{An Example of Feedback.}
    In a user behavior analysis application, each user visit is depicted as an attributed sequence. Given two user sessions $p_1$ and $p_2$, domain experts may imply they are \textit{similar} by giving the feedback $(p_1, p_2, 0)$; or imply they are \textit{dissimilar} by the feedback $(p_1, p_2, 1)$. 

\subsection{Problem definition} Given a nonlinear transformation function $\Theta$ and two attributed sequences $p_i$ and $p_j$ as inputs, deep metric learning approaches~\cite{cvpr-face-verify} often apply the Mahalanobis distance function to the $d$-dimensional outputs of function $\Theta$ as: %
\begin{equation}
\label{eq-mahalanobis-distance}
    D_\Theta(p_i, p_j) = \sqrt{(\Theta(p_i) - \Theta(p_j))^\top \pmb\Lambda (\Theta(p_i) - \Theta(p_j))}
\end{equation}
where $\pmb\Lambda \in \mathbb{R}^{d \times d}$ is a symmetric, semi-definite, and positive matrix. When $\pmb\Lambda = \mathbf{I}$, Equation~\ref{eq-mahalanobis-distance} is equivalent to Euclidean distance~\cite{xing2003distance} as:
\begin{equation}
\label{eq-euclidean}
    D_\Theta(p_i, p_j) = \|\Theta(p_i)-\Theta(p_j)\|_2.
\end{equation}
 
Given feedback sets $\mathcal{S}$ and $\mathcal{D}$ of attributed sequences as per Def.~\ref{def-as-feedback} and a distance function $D_\Theta$ as per Equation~\ref{eq-euclidean}, the goal of deep metric learning on attributed sequences is to find the transformation function $\Theta: (\mathbb{R}^u, \mathbb{R}^{T \times r}) \mapsto \mathbb{R}^d$ with a set of parameters $\theta$ that is capable of mapping the attributed sequence inputs to a space that the distances between each pair of attributed sequences in the similar feedback set $\mathcal{S}$ are minimized while increasing the distances between attributed sequence pairs in the dissimilar feedback set $\mathcal{D}$. Inspired by~\cite{xing2003distance}, we adopt the learning objective as:
\begin{equation}
\label{eq-problem-def}
\begin{split}
\minimize_{\theta} &\sum_{(p_i, p_j, \ell_{ij})\in \mathcal{S}} D_\Theta\left(p_i, p_j\right) \\[2pt]
\textrm{s.t.} & \sum_{(p_i, p_j, \ell_{ij})\in \mathcal{D}} D_\Theta\left(p_i, p_j\right) \geq g\\
\end{split}
\end{equation}
where $g$ is a group-based margin parameter that stipulates the distance between two attributed sequences from dissimilar feedback set should be larger than $g$. This prevents the dataset from being reduced to a single point~\cite{xing2003distance}.  %
\section{The \sysname Network Architecture}
\label{section-model}
We first design two networks to handle attributed sequence data. Namely, the \anet using feedforward fully connected neural networks for the attributes and the \snet using long short-term memory (LSTM) for the sequences. Next, we explore three design variations of the \fnet concerning the attribute-sequence dependencies. Lastly, the \fnet is augmented by \mnet to exploit the user feedback. 

\subsection{\anet and \snet Design} 
\label{subsection-att-seq-net}
\subsubsection{\anetns}
\anet is designed using a fully connected neural network to learn the relationships within attribute data. In particular, for an \anet with $M$ layers, we denote the weight and bias parameters of the $m$-th layer as $\mathbf{W}_A^{(m)}$ and $\mathbf{b}_A^{(m)}, \forall m=1,\cdots, M$. Given an attribute vector $\mathbf{x}_k \in \mathbb{R}^u$ as the input, with $d_m$ hidden units in the $m$-th layer of \anetns, the corresponding output is $\mathbf{V}^{(m)}\in\mathbb{R}^{d_m}, \forall m=1,\cdots, M$. Note that the choice of $M$ is task-specific. Although neural networks with more layers are better at learning hierarchical structure in the data, it is also observed that such networks are difficult to train due to the multiple nonlinear mappings that prevent the information and gradient passing along the computation graph~\cite{pham2017column}. The \anet is designed as: 
\begin{equation}
\small
\begin{split}
\label{eq-att-network}
\mathbf{V}^{(1)} &= \delta\left(\mathbf{W}_A^{(1)}\mathbf{x}_k + \mathbf{b}_A^{(1)}\right) \\[-6pt]
\vdots \\[-3pt]
\mathbf{V}^{(M)} &= \delta\left(\mathbf{W}_A^{(M)}\mathbf{V}^{(M - 1)} + \mathbf{b}_A^{(M)}\right)
\end{split}
\end{equation}
where $\delta: \mathbb{R}^{d_m}\mapsto\mathbb{R}^{d_m}$ is a hyperbolic tangent activation function. %

Given the attribute vector $\mathbf{x}_k \in \mathbb{R}^u$ as the input, the first layer of \anet uses the weight matrix $\mathbf{W}_A^{(1)} \in \mathbb{R}^{d_1 \times u}$ and bias vector $\mathbf{b}_A^{(1)}\in\mathbb{R}^{d_1}$ to map $\mathbf{x}_k$ to the output $\mathbf{V}^{(1)} \in \mathbb{R}^{d_1}$ with $d_1<u$. 
The $\mathbf{V}^{(1)}$ is subsequently used as the input to the next layer. 
We denote the \anet as 
$\Theta_A: \mathbb{R}^{u} \mapsto \mathbb{R}^{d_{M}}$, where
$\Theta_A$ is parameterized by $\mathbf{W}_A$ and $\mathbf{b}_A$, where $\mathbf{W}_A = \left[\mathbf{W}_A^{(1)}, \cdots, \mathbf{W}_A^{(M)}\right]$ and $\mathbf{b}_A = \left[\mathbf{b}_A^{(1)}, \cdots, \mathbf{b}_A^{(M)}\right]$.

\subsubsection{\snetns}

The \snet is designed to learn the dependencies between items in the input sequences by taking advantage of the long short-term memory (LSTM)~\cite{hochreiter1997long} network to learn the dependencies within the sequences.  

\begin{figure}[t]
    \centering
    \begin{subfigure}{0.45\linewidth}
    \centering
        \includegraphics[width=1\textwidth]{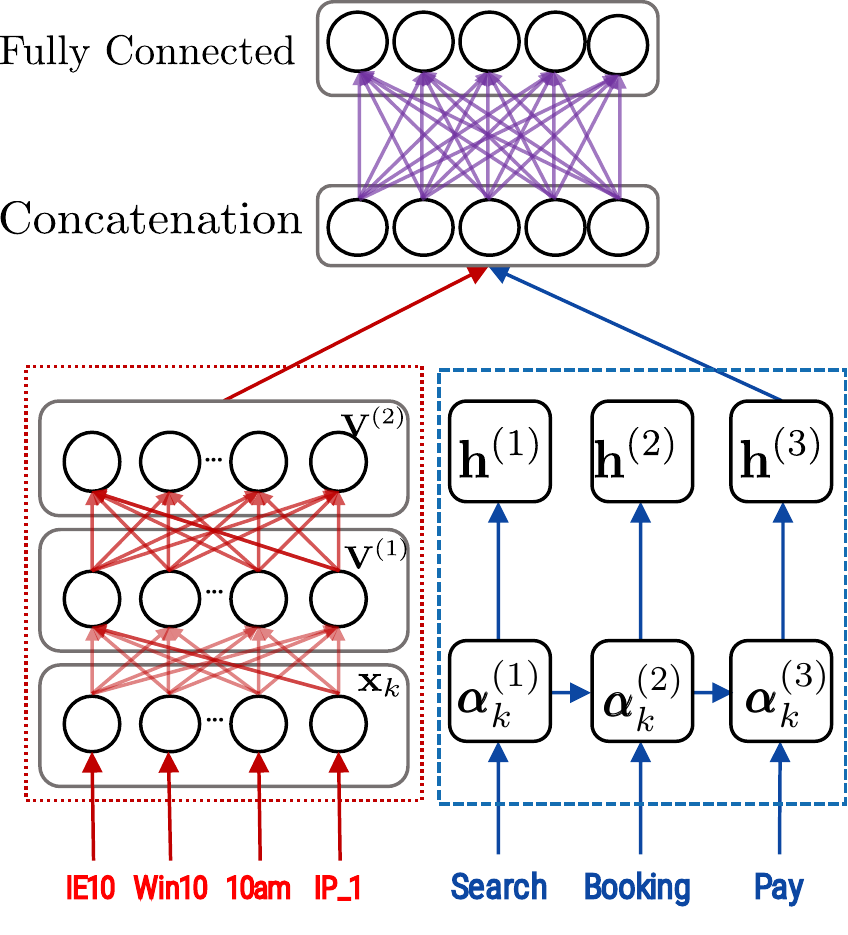}
        \captionsetup{format=centercaption}
        \caption{Balanced Design}%
        \label{fig-balanced}
    \end{subfigure}
    \begin{subfigure}{0.45\linewidth}
        \begin{subfigure}{\linewidth}
            \centering
            \includegraphics[width=1\textwidth]{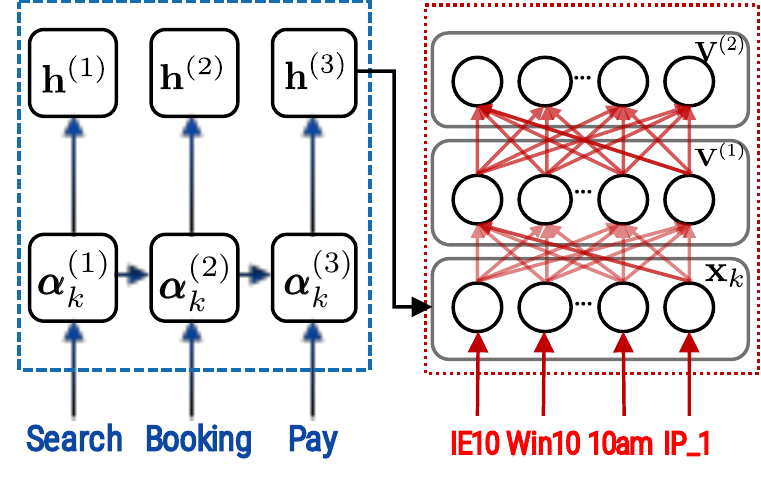}
            \vspace{-5mm}
            \caption{\anetns-centric Design}%
            \label{fig-att-centric}
        \end{subfigure}
        
        \begin{subfigure}{\linewidth}
            \centering
            \includegraphics[width=1\textwidth]{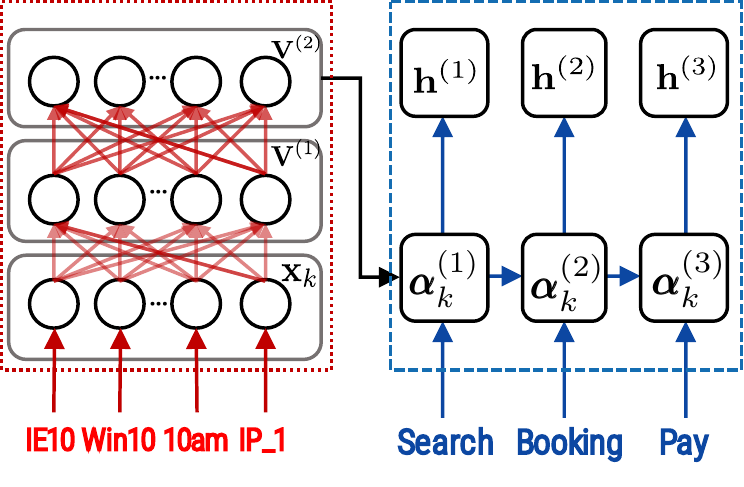}
            \vspace{-5mm}
            \caption{\snetns-centric Design} %
            \label{fig-seq-centric}
            \vspace{-3mm}
        \end{subfigure}
    \end{subfigure}
    \caption{Three different \fnet designs. }%
    \vspace{-5mm}
    \label{fig-three-designs}
\end{figure}
 
Given a sequence input $\mathbf{S}_k \in \mathbb{R}^{T\times r}$, we have the parameters within \snet at time step $t$ as:
\begin{equation}
  \begin{split}
  \label{eq-lstm}
  \mathbf{i}^{(t)} &= \sigma\left(\mathbf{W}_{i}\vec{\alpha}_k^{(t)} + \mathbf{U}_{i}\mathbf{h}^{(t-1)} + \mathbf{b}_i\right) \\[-2pt]
  \mathbf{f}^{(t)} &= \sigma\left(\mathbf{W}_{f}\vec{\alpha}_k^{(t)} + \mathbf{U}_{f}\mathbf{h}^{(t-1)} + \mathbf{b}_f\right) \\[-2pt]
  \mathbf{o}^{(t)} &= \sigma\left(\mathbf{W}_{o}\vec{\alpha}_k^{(t)} + \mathbf{U}_{o}\mathbf{h}^{(t-1)} + \mathbf{b}_o\right) \\[-2pt]
  \mathbf{g}^{(t)} &= \tanh\left(\mathbf{W}_{c}\vec{\alpha}_k^{(t)} + \mathbf{U}_{c}\mathbf{h}^{(t-1)} + \mathbf{b}_c\right) \\[-2pt]
  \mathbf{c}^{(t)} &= \mathbf{f}^{(t)}\odot\mathbf{c}^{(t-1)} + \mathbf{i}^{(t)} \odot \mathbf{g}^{(t)} \\[-2pt]
  \mathbf{h}^{(t)} &= \mathbf{o}^{(t)} \odot \tanh\left(\mathbf{c}^{(t)}\right)
  \end{split}
\end{equation}
where $\sigma(z)=\frac{1}{1+e^{-z}}$ is a logistic activation function, $\odot$ denotes the bitwise multiplication, $\mathbf{i}^{(t)}$, $\mathbf{f}^{(t)}$ and $\mathbf{o}^{(t)}$ are the internal gates of the LSTM, $\mathbf{c}^{(t)}$ and $\mathbf{h}^{(t)}$ are the cell and hidden states of the LSTM. 
For simplicity, we denote the \snet with $d_S$ hidden units as  
$\Theta_S:\mathbb{R}^{T \times r} \mapsto \mathbb{R}^{d_S}$, 
where $\Theta_S$ is parameterized by bias vector set $\mathbf{b}_S=\left[\mathbf{b}_i,\mathbf{b}_f,\mathbf{b}_o,\mathbf{b}_c\right]$ and the weight matrices $\mathbf{W}_{S}\!=\!\left[\mathbf{W}_{i}, \mathbf{W}_{f}, \mathbf{W}_{o}, \mathbf{W}_{c}\right]\in \mathbb{R}^{4\times d_S \times r}$ and recurrent matrices $\mathbf{U}_{S} = \left[\mathbf{U}_{i}, \mathbf{U}_{f}, \mathbf{U}_{o}, \mathbf{U}_{c}\right]\in \mathbb{R}^{4\times d_S \times d_S}$. 

\subsection{\fnet Design}
\label{subsection-fusion-net}

One important design in \sysname is the \fnet (denoted as $\Theta$), where the \anet and \snet are connected together with the goal of producing feature representation outputs for the attributed sequences. 
Based on the possible ways of connecting the \anet and the \snetns, we propose three \fnet design variations: (1) balanced, (2) \anetns-centric and (3) \snetns-centric. We evaluate the performance of all three variations in Section~\ref{section-experiments}.

\subsubsection{Balanced Design (Figure~\ref{fig-balanced}).} 
\label{sec-design-balance} 
The \anet and \snet are concatenated together and augmented by an additional layer fully connected neural network. This additional layer of fully connected neural network is used to capture the dependencies between the attributes and the sequences. For each attributed sequence, we only use the output of \snet after the last time step of the sequence has been processed to capture the complete information of the sequence. We denote the balanced design as:  
\begin{align*}
    \normalsize
    \mathbf{y} &= \mathbf{V}^{(M)}\oplus \mathbf{h}^{(T_k)}\\
    \mathbf{z} &= \delta\left(\mathbf{W}_z\mathbf{y} + \mathbf{b}_z\right)
\end{align*}
where $\oplus$ represents the concatenation operation, $\mathbf{W}_z\in\mathbb{R}^{d\times(d_M+d_S)}$ and $\mathbf{b}_z\in\mathbb{R}^{d}$ denote the weight matrix and bias vector in this fully connected layer, respectively. 
\begin{figure*}[t]
    \centering
        \includegraphics[width=0.5\linewidth]{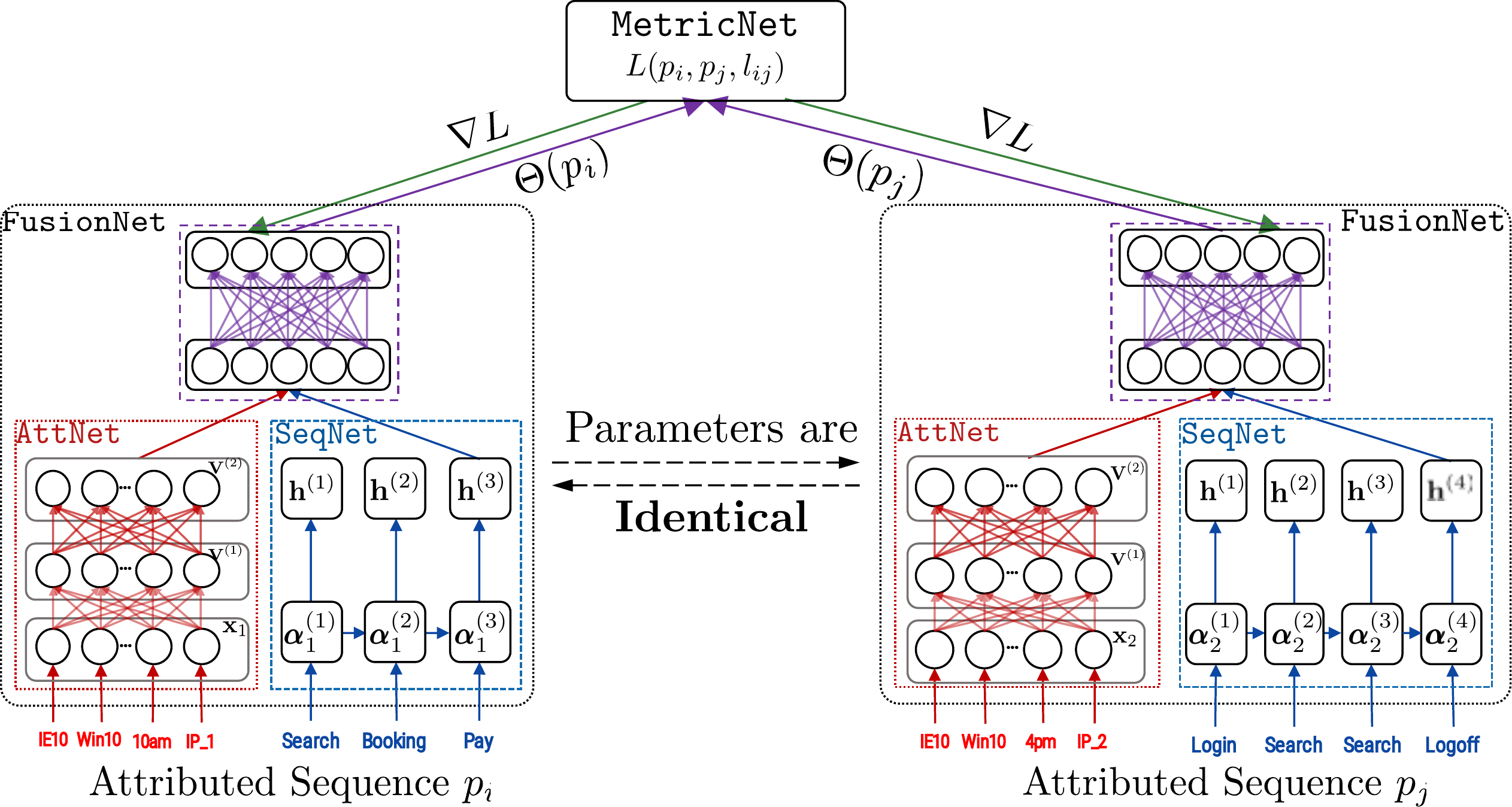} 
        \caption{\sysname network with balanced design. Parameters in the two \fnet are shared. The feature representations are used by \mnet to compute the loss $L$. The gradient $\nabla L$ is used to update all layers. }%
        \label{fig-fdml}
\end{figure*} %
\subsubsection{\anetns-centric Design (Figure~\ref{fig-att-centric}).}
\label{sec-design-attribute}
Here, the sequence is first transformed by the sequence network, \ie, the function $\Theta_S$, and then used as an input of the \anetns. Specifically, we modify Equation~\ref{eq-att-network} to incorporate sequence representation as input. Similar to the balanced design, only the output of \snet after processing the last time step of the sequence is used to capture the complete information. The modified $\mathbf{V}_k^{(1)}$ is written as:  
\begin{equation*}
    \mathbf{V}^{(1)} = \delta\left(\mathbf{W}_A^{(1)}\left(\mathbf{x_k}\oplus\mathbf{h}^{(T_k)}\right)+\mathbf{b}_A^{(1)}\right)
\end{equation*}
where the $\mathbf{W}_A^{(1)}\in \mathbb{R}^{d_1\times(u+d_S)}$ and $\mathbf{b}_A^{(1)}\in\mathbb{R}^{d_1}$. 

\subsubsection{\snetns-centric Design (Figure~\ref{fig-seq-centric}).}
\label{sec-design-sequence}
The output of \anet is used as an additional input for \snet. Specifically, we modify Equation~\ref{eq-lstm} to integrate attribute representations at the first time step. The modified $\mathbf{h}_k^{(t)}$ is defined as:  
\begin{equation*}
\mathbf{h}^{(1)} = \mathbf{o}^{(1)} \odot \tanh\left(\mathbf{c}^{(1)}\right) + \mathbf{V}^{(M)}
\end{equation*}
where $\odot$ denotes the bitwise multiplication. The \snet is conditioned using the information from \anetns. 

\subsection{The \mnetns}
\label{subsection-metric-net}
We present the \mnet using the proposed balanced design due to space limitations. In the balanced design (as shown in Figure~\ref{fig-balanced}), the explicit form of \fnet can be written as: 
\begin{equation}
\label{eq-balanced-specific}
    \Theta(p_k) = \Theta_A\left(\Theta_A(\mathbf{x}_k)\oplus\Theta_S(\mathbf{S}_k)\right)
\end{equation}
    
Given an attributed sequence feedback instance $(p_i, p_j, \ell_{ij})$, where $p_i = (\mathbf{x}_i, \mathbf{S}_i)$ and $p_j = (\mathbf{x}_j, \mathbf{S}_j)$, $\ell_{ij} \in \{1, 0\}$. This input is transformed to $\Theta(p_i) \in \mathbb{R}^d$ and $\Theta(p_j)\in \mathbb{R}^d$ by the nonlinear transformation $\Theta$ of \fnet at the first step. 

Next, the two outputs of \fnet are taken by the \mnet to calculate the differences between them. 
The \mnet is designed to utilize a contrastive loss function~\cite{hadsell2006dimensionality} so that attributed sequences in each similar pair in $\mathcal{S}$ have a smaller distance compared to those in $\mathcal{D}$ after learning the distance metric. The learning objective of \mnet can be written as:
\begin{equation}
\label{eq-contrastive-loss} 
L(p_i, p_j, \ell_{ij}) = \frac{1}{2}(1-\ell_{ij})(D_\Theta)^2 + \frac{1}{2}\ell_{ij}\{\max(0, g - D_\Theta)\}^2
\end{equation}
where $g$ is the margin parameter, meaning that the pairs with a dissimilar label ($\ell_{ij}=1$) contribute to the learning objective if and only if when the Euclidean distance between them is smaller than $g$~\cite{hadsell2006dimensionality}. 

For each pair of \fnet outputs, \mnet first computes the Euclidean distance (Equation~\ref{eq-euclidean}) between them. Then the contrastive loss is computed using both the Euclidean distance and the label. 
We note that the \mnet can augment all three designs in the same way. Figure~\ref{fig-fdml} illustrates the \mnet with the proposed balanced design. 

The parameters in both \fnet and \mnet are adjusted using the metric learning presented in Algorithm~\ref{alg-fase}. 
\begin{algorithm}[t]
    \footnotesize
    \begin{algorithmic}[1]
    \caption{Training the \sysname Network}
    \label{alg-fase}
    \normalsize
    \INPUT A set of attributed sequences $\mathcal{P} = \{p_1, \cdots, p_n\}$, a set of feedback as pairwise attributed sequences $\mathcal{C} = \{(p_i, p_j, \ell_{ij})|p_i, p_j \in \mathcal{P},
    \}$, the number of layers $M$, learning rate $\gamma$, number of iterations $\vartheta$ and convergence error $\epsilon$. 
    \OUTPUT Parameter sets $\{\mathbf{W}_A, \mathbf{b}_A, \mathbf{W}_S, \mathbf{U}_S, \mathbf{b}_S\}$.
    \State{Initialize \sysname network $\Theta$.}
    \State{Pre-train the \sysname network.}
    \ForEach{$\vartheta^{\prime} = 1, \cdots, \vartheta$}
        \ForEach{$(p_i, p_j, \ell_{ij}) \in \mathcal{C}$}
            \State{//Forward propagation.}
            \State{Calculate $\Theta(p_i)$ and $\Theta(p_j)$.}
            \State{Calculate $D_\Theta$ using Equation~\ref{eq-euclidean}.}
            \State{Calculate loss $L_{\vartheta^{\prime}}(p_i, p_j, \ell_{ij})$ (Equation~\ref{eq-contrastive-loss}).}
            \If{$|L_{\vartheta^{\prime}}(p_i, p_j, \ell_{ij}) - L_{\vartheta^{\prime}-1}(p_i, p_j, \ell_{ij})| < \epsilon$}
                \State{\textbf{break}}
            \EndIf
            \State{//Back-propagation.}
            \State{Calculate $\frac{\partial L}{\partial \Theta}$.}
            \State{Calculate $\nabla L$.}
            \State{Update network parameters.}
        \EndFor
    \EndFor
    \end{algorithmic}
\end{algorithm}

\section{Experiments}
\label{section-experiments}
\subsection{Datasets}
We evaluate the proposed methods using four real-world datasets. Two of them are derived from application log files\footnote{Personal identity information is not collected. } at Amadeus~\cite{amadeus} (denoted as AMS-A and AMS-B). The other two datasets are derived from the Wikispeedia~\cite{west2009wikispeedia} dataset (denoted as Wiki-A and Wiki-B). For each dataset, we randomly sampled 80\% as the training set and 20\% as the testing set. The training and testing sets remain the same across all our experiments. 
\begin{table}[t]
  \caption{Summary of Compared Methods}
  \label{table-compared-methods}
  \small
  \centering
  \begin{tabular}{ccp{1.25in}p{1.3cm}}
    \toprule
    Method & \thead{Data Used} & \thead{Short Description} & \thead{Reference} 
    \\ \midrule
    \multirow{2}{*}{\texttt{ATT}} & \multirow{2}{*}{Attributes} & \multirow{2}{*}{\shortstack[l]{Only \textit{attribute} feedback \\ is used in the model.}} & \multirow{2}{*}{\cite{cvpr-face-verify}}
    \\ & & &
    \\ \midrule
    \multirow{2}{*}{\texttt{SEQ}} & \multirow{2}{*}{Sequences} & \multirow{2}{*}{\shortstack[l]{Only \textit{sequence} feedback \\ is used in the model.}}& \multirow{2}{*}{\cite{mueller2016siamese}}
    \\ & & &
    \\ \midrule
    \multirow{3}{*}{\texttt{ASF}} & \multirow{3}{*}{\shortstack{Attributes \\ Sequences}} & \multirow{3}{*}{\shortstack[l]{Feedback of attributes and \\sequences are used to train \\ two models \textbf{\textit{separatedly}}. }} & \multirow{3}{*}{\cite{cvpr-face-verify} + \cite{mueller2016siamese}}
    \\ & & &
    \\ & & &
    \\ \midrule
    \multirow{3}{*}{\sysnamenospace\texttt{-B}} & \multirow{3}{*}{\shortstack{Attributes \\ Sequences \\ Dependencies}} &\multirow{3}{*}{\shortstack[l]{Balanced design using attri- \\ -buted sequence feedback \\ to train one \textbf{\textit{unified}} model. }} & \multirow{3}{*}{This Work} 
    \\ & & &
    \\ & & &
    \\ \midrule
    \multirow{4}{*}{\sysnamenospace\texttt{-A}} & \multirow{4}{*}{\shortstack{Attributes \\ Sequences \\ Dependencies}} & \multirow{4}{*}{\shortstack[l]{Attribute-centric design \\ using attributed sequence \\ feedback to train one \\ \textbf{\textit{unified}} model. }} & \multirow{4}{*}{This Work} 
    \\ & & &
    \\ & & &
    \\ & & &
    \\ \midrule
    \multirow{4}{*}{\sysnamenospace\texttt{-S}} & \multirow{4}{*}{\shortstack{Attributes \\ Sequences \\ Dependencies}} & \multirow{4}{*}{\shortstack[l]{Sequence-centric design \\ using attributed sequence \\ feedback to train one \\ \textbf{\textit{unified}} model. }} & \multirow{4}{*}{This Work} 
    \\ & & &
    \\ & & &
    \\ & & &    
    \\ \bottomrule
  \end{tabular}
  \vspace{-5mm}
\end{table}
 
\begin{itemize}
    \item \textbf{AMS-A}: We used 58k user sessions from log files of an internal application from our Amadeus partner company. Each record has a user profile containing information ranging from system configurations to office name, and a sequence of functions invoked by click activities on the web interface. There are 288 distinct functions, 57,270 distinct user profiles in this dataset. The average length of the sequences is 18. We use 100 attributed sequence feedback pairs selected by the domain experts. 
    \item \textbf{AMS-B}: There are 106k user sessions derived from application log files with 575 distinct functions and 106,671 distinct user profile. The average length of the sequences is 22. Domain experts select 84 attributed sequence feedback pairs. 
    \item \textbf{Wiki-A}: This dataset is sampled from Wikispeedia dataset. Wikispeedia dataset originated from an online computation game~\cite{west2009wikispeedia}, in which each user is given two pages ({\ie}, source, and destination) from a subset of Wikipedia pages and asked to navigate from the source to the destination page. We use a subset of $\sim$2k paths from Wikispeedia with the average length of the path as 6. We also extract 11 sequence context as attributes (\eg, the category of the source page, average time spent on each page, \etc). There are 200 feedback instances selected based on the criteria of frequent subsequences and attribute value. 
    \item \textbf{Wiki-B}: This dataset is also sampled from Wikispeedia dataset. We use a subset of $\sim$1.5k paths from Wikispeedia with the average length of the path as 8. We also extract 11 sequence context (\eg, the category of the source page, average time spent on each page, \etc) as attributes. 220 feedback instances have been selected based on the criteria of frequent sub-sequences and attribute values.\end{itemize}

\subsection{Compared Methods} 

We validate the effectiveness of our proposed \sysname solution compared with state-of-the-art baseline methods. To well understand the advancements of the proposed methods, we use baselines that are working on only attributes (denoted as \texttt{ATT}) or sequences (denoted as \texttt{SEQ}), as well as methods without exploiting the dependencies between attributes and sequences (denoted as \texttt{ASF}) . %
 
1. \underline{Att}ribute-only Feedback (\texttt{ATT})~\cite{cvpr-face-verify}: Only attribute feedback is used in this model. This model first transforms fixed-size input data into feature vectors, then learns the similarities between these two inputs. 

2. \underline{Seq}uence-only Feedback (\texttt{SEQ})~\cite{mueller2016siamese}: Only sequence feedback is used in this model. This model utilizes a long short-term memory (LSTM) to learn the similarities between two sequences.

3. \underline{A}ttribute and \underline{S}equence \underline{F}eedback (\texttt{ASF})~\cite{cvpr-face-verify} + \cite{mueller2016siamese}: This method is a combination of the \texttt{ATT} and \texttt{SEQ} methods as above, where the two networks are trained \textit{separately} using attribute feedback and sequence feedback, respectively. The feature vectors generated by the two models are then concatenated. 

4. Balanced Network Design with Attributed Sequence Feedback (\sysnamenospace\texttt{-B}): This is the balanced design model (Section~\ref{sec-design-balance}) using attributed sequence feedback to train one {\textit{unified}} model.
  
5. \anetns-centric Network Design with Attributed Sequence Feedback (\sysnamenospace\texttt{-A}): This is the \anetns-centric design (Section~\ref{sec-design-attribute}) using attributed sequence feedback to train one {\textit{unified}} model. 

6. \snetns-centric Network Design with Attributed Sequence Feedback (\sysnamenospace\texttt{-S}): This is the \snetns-centric design (Section~\ref{sec-design-sequence}) using attributed sequence feedback to train one {\textit{unified}} model.
\begin{figure*}[t]
    \centering
    \begin{subfigure}[t]{\figwidth\linewidth}
      \centering
      \includegraphics[page=1, width=1.05\linewidth]{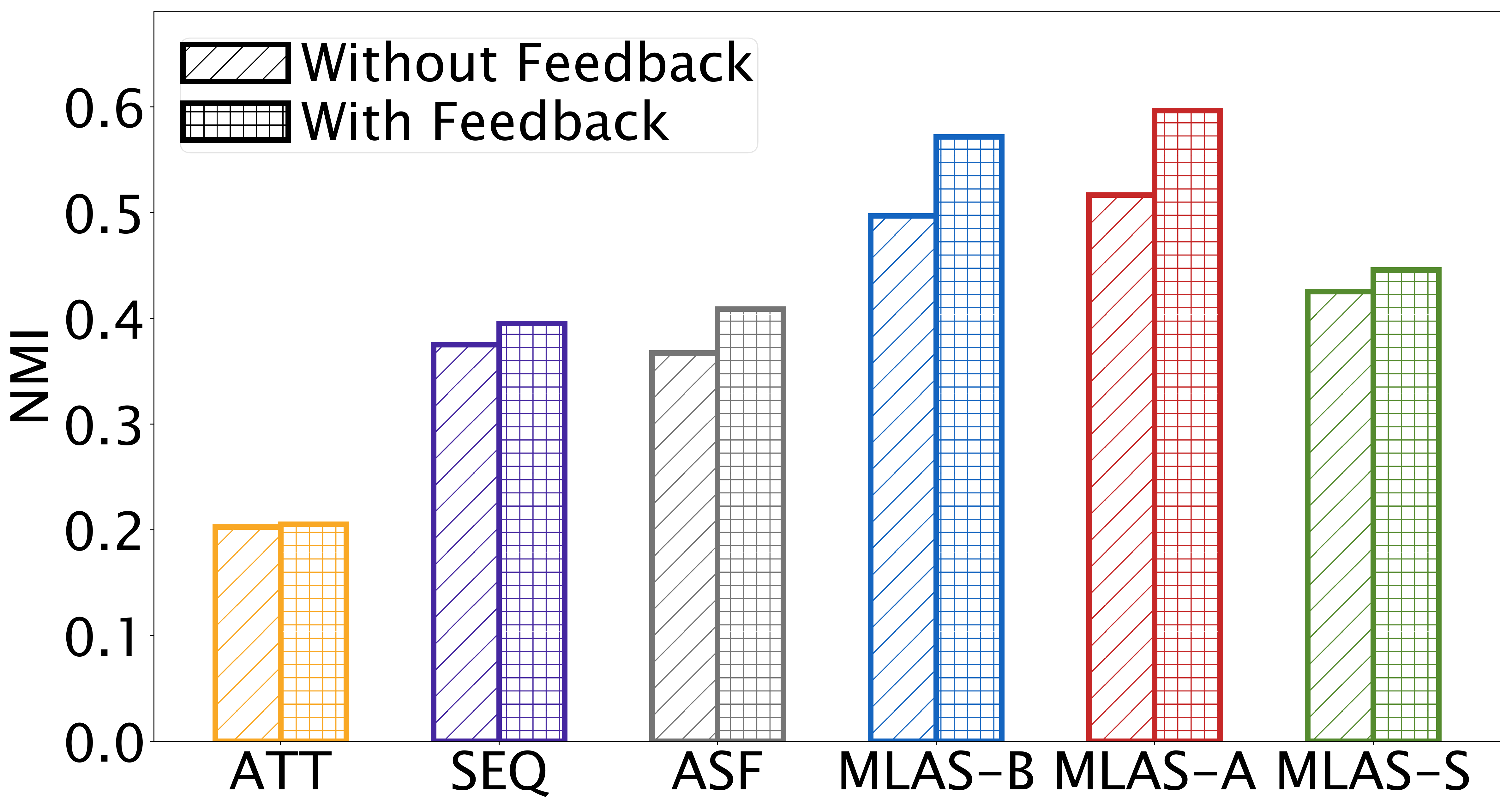}
      \label{fig-exp-fb1}      
      \vspace{-5mm}
      \caption{AMS-A Dataset}
    \end{subfigure}
    \hfill
    \begin{subfigure}[t]{\figwidth\linewidth}
      \centering
      \includegraphics[page=2, width=1.05\linewidth]{nmi_dim_bar}
      \label{fig-exp-fb2}
      \vspace{-5mm}
      \caption{AMS-B Dataset}
    \end{subfigure}
    \hfill
    \begin{subfigure}[t]{\figwidth\linewidth}
      \centering
      \includegraphics[page=3, width=1.05\linewidth]{nmi_dim_bar}
      \label{fig-exp-fb3}
      \vspace{-5mm}
      \caption{Wiki-A Dataset}
    \end{subfigure}
    \hfill
    \begin{subfigure}[t]{\figwidth\linewidth}
      \centering
      \includegraphics[page=4, width=1.05\linewidth]{nmi_dim_bar}
      \label{fig-exp-fb4}
      \vspace{-5mm}
      \caption{Wiki-B Dataset}
    \end{subfigure}
    \vspace{-3mm}
  \caption{The effectiveness of using feedback. Using feedback could boost performance of all methods. The three methods we proposed (\sysnamenospace\texttt{-B/A/S}) are capable of exploiting the information of attributes, sequences, and more importantly, the attribute-sequence dependencies to outperform other methods. }
  \label{fig-exp-feedback}
  \vspace{-3mm}
\end{figure*}
\begin{figure*}[t]
  \centering
  \includegraphics[width=0.5\linewidth]{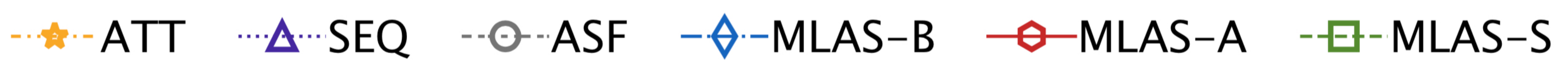}
  \vspace{-1mm}

    \begin{subfigure}[t]{0.24\linewidth}
      \centering
      \includegraphics[page=1, width=\linewidth]{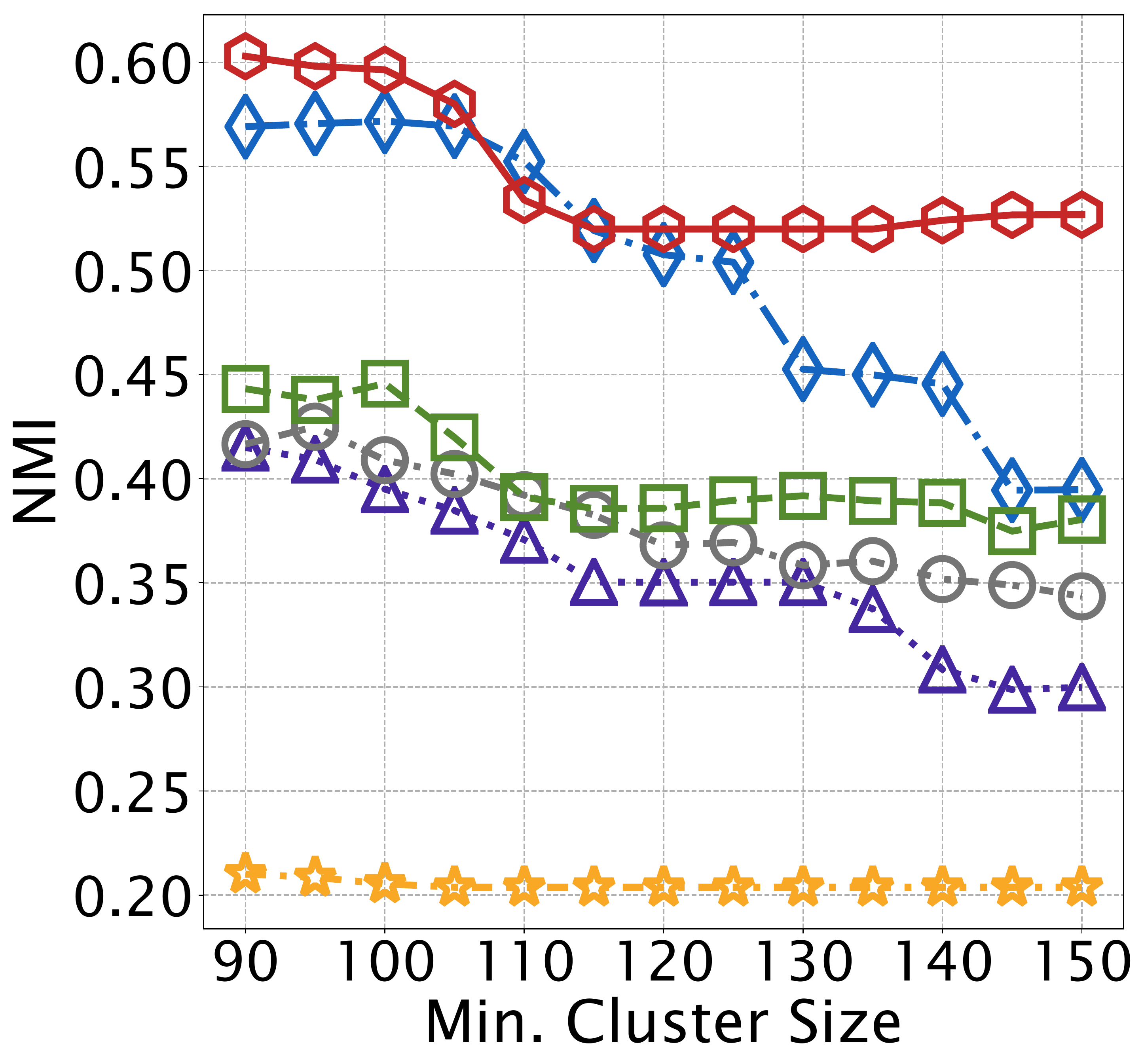}
      \label{fig-exp-cs1}
      \vspace{-5mm}
      \caption{AMS-A Dataset}
    \end{subfigure}
    \hfill
    \begin{subfigure}[t]{0.24\linewidth}
      \centering
      \includegraphics[page=2, width=\linewidth]{nmi_cs}
      \label{fig-exp-cs2}
      \vspace{-5mm}
      \caption{AMS-B Dataset}
    \end{subfigure}
    \hfill
    \begin{subfigure}[t]{\figwidth\linewidth}
      \centering
      \includegraphics[page=3, width=\linewidth]{nmi_cs}
      \label{fig-exp-cs3}
      \vspace{-5mm}
      \caption{Wiki-A Dataset}
    \end{subfigure}
    \hfill
    \begin{subfigure}[t]{\figwidth\linewidth}
      \centering
      \includegraphics[page=4, width=\linewidth]{nmi_cs}
      \label{fig-exp-cs4}
      \vspace{-5mm}
      \caption{Wiki-B Dataset}
    \end{subfigure}
    \vspace{-3mm}
  \caption{Performance with varying clustering parameters. Clustering results using the feature representations produced by \sysname are the best among the compared methods. }%
  \label{fig-exp-clustering}
  \vspace{-5mm}
\end{figure*}
 
\subsection{Experimental Settings} In this section, we present the settings for performance evaluation and parameter studies. 
\subsubsection{Network Initialization and Training} Initializing the network parameters is important for models using gradient descent based approaches~\cite{erhan2009difficulty}. The weight matrices $\mathbf{W}_A$ in $\Theta_A$ and the $\mathbf{W}_{S}$ in $\Theta_S$ are initialized using the uniform distribution~\cite{glorot2010understanding}, the biases $\mathbf{b}_A$ and $\mathbf{b}_S$ are initialized with zero vector $\pmb0$ and the recurrent matrix $\mathbf{U}_{S}$ is initialized using orthogonal matrix~\cite{saxe2013exact}. We use one hidden layer ($M=1$) for \anet and \texttt{ATT} in the experiments to make the training process easier.  %

After that, we pre-train each compared method. Pre-training is an important step to initialize the neural network-based models~\cite{erhan2009difficulty}. Our pre-training uses the attributed sequences as the inputs for \fnetns, and use the generated feature representations to reconstruct the attributed sequence inputs. We also pre-train the \texttt{ATT} and \texttt{SEQ} networks in a similar fashion that reconstruct the respective attributes or sequences. 
We utilize $\ell_2$-regularization and early stopping strategy to avoid overfitting. Twenty percents of feedback pairs are used in the validation set. We choose \texttt{ReLU} activation function~\cite{nair2010rectified} in our \anet to accelerate the stochastic gradient descent convergence.

\subsubsection{Performance Evaluation Setting.}
We evaluate the performance by using the feature representations generated by each method for clustering tasks. The feature representations are generated through a forward pass. 

Clustering tasks have been widely used in distance metric learning work~\cite{xing2003distance,wang2015survey}. In this set of experiments, we use HDBSCAN~\cite{campello2015hierarchical} clustering algorithm. HDBSCAN is a deterministic algorithm, which gives identical output when using the same input. We measure the normalized mutual information (NMI)~\cite{mcdaid2011normalized} score. The maximum NMI score is 1. Specifically, we conduct the below two experiments:

1. The effect of feedback. We compare the performance of the clustering algorithm using the feature representations generated by \fnet before and after incorporating the feedback. 

2. The effect of varying parameters in the clustering algorithm. After the metric learning process, we evaluate the feature representations generated by all compared methods under various parameters of the clustering algorithm. 

\subsubsection{Parameter Study Settings.} 
We evaluate the effect of output dimension (\ie, the dimension of the hidden layer), which affects the model size and the performance of downstream mining algorithms. 

The other parameter we evaluate is the relative importance of attribute data (denoted as $\omega_A$) in the attributed sequences. The pre-training phase is essential to gradient descent-based methods~\cite{erhan2009difficulty}. The relative importance of attribute data and sequence data are represented by the weights of $\Theta_A$ and $\Theta_S$, denoted as $\omega_A$ and $\omega_S$, where $\omega_A + \omega_S = 1$. The intuition is that with one data type more important, the other one becomes relatively less important. 

\subsection{Results and Analysis}

\subsubsection{Effect of Feedback.} We present the performance comparisons in clustering tasks using feature representations generated using the parameters of all methods in Figure~\ref{fig-exp-feedback}. Two sets of feature representations are generated, the first set is generated after the pre-training (denoted as \textit{without feedback}), the other set is generated after the metric learning step (denoted as \textit{with feedback}). We fix the output dimension to 10, minimum cluster size to 100 and $\omega_A=0.5$ (for \sysnamenospace\texttt{-B/A/S}). We have observed that the feedback can boost the performance of all methods, and the three methods (\sysnamenospace\texttt{-B/A/S}) proposed in this work are capable of outperforming other methods. Also, we also observe that the proposed three \sysname variations have better performance compared to the \texttt{ASF}, which also uses the information from attributes and sequences but \textit{without} using the attribute-sequence dependencies. 

Based on the above observations, we can conclude that the performance boost of our three architectures (\sysnamenospace\texttt{-B/A/S}) is a result of taking advantages of attribute data, sequence data, and more importantly, the attribute-sequence dependencies. 
\begin{figure*}[t]
  \centering
  \includegraphics[width=0.5\linewidth]{legend-6}
  \vspace{-1mm}
  
      \begin{subfigure}[t]{\figwidth\linewidth}
        \centering
        \includegraphics[page=1, width=\linewidth]{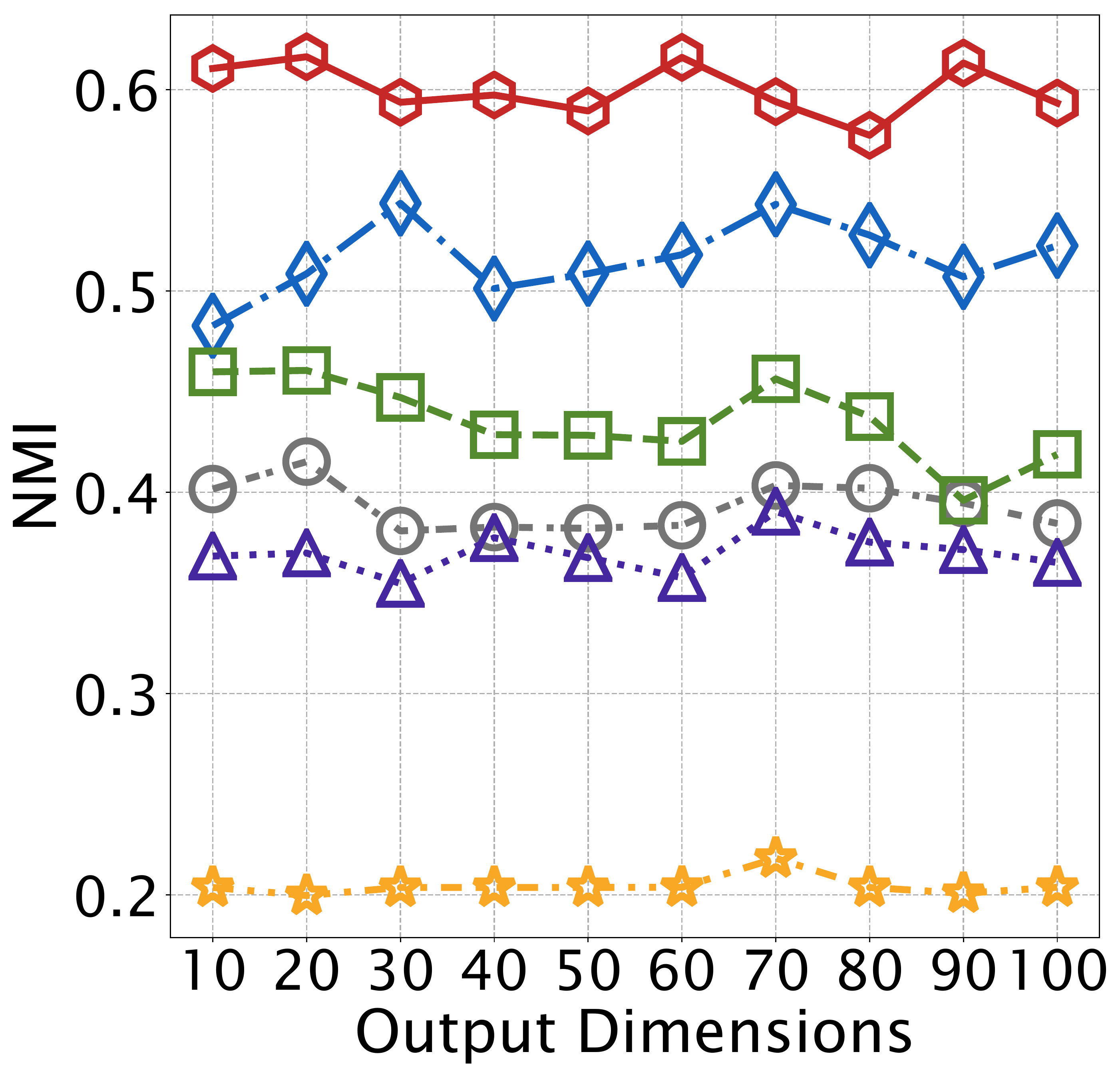}
        \label{fig-exp-dim1}
        \vspace{-5mm}
        \caption{AMS-A Dataset}
      \end{subfigure}
      \hfill
      \begin{subfigure}[t]{0.24\linewidth}
        \centering
        \includegraphics[page=2, width=\linewidth]{nmi_dim}
        \label{fig-exp-dim2}
        \vspace{-5mm}
        \caption{AMS-B Dataset}
      \end{subfigure}
      \hfill
      \begin{subfigure}[t]{\figwidth\linewidth}
        \centering
        \includegraphics[page=3, width=\linewidth]{nmi_dim}
        \label{fig-exp-dim3}
        \vspace{-5mm}
        \caption{Wiki-A Dataset}
      \end{subfigure}
      \hfill
      \begin{subfigure}[t]{0.24\linewidth}
        \centering
        \includegraphics[page=4, width=\linewidth]{nmi_dim}
        \label{fig-exp-dim4}
        \vspace{-5mm}
        \caption{Wiki-B Dataset}
      \end{subfigure}
      \vspace{-3mm}
     \caption{The effect of output dimensions (higher is better). Output dimension is an important factor for (1) the size of model; and (2) the cost of computations in downstream data mining tasks. Using the feature representations produced by \sysname can  constantly achieve the best performance among the compared methods. }
    \label{fig-exp-dimension}
    \vspace{-3mm}
  \end{figure*}
\begin{figure*}[t]
  \centering
  \includegraphics[width=0.3\linewidth]{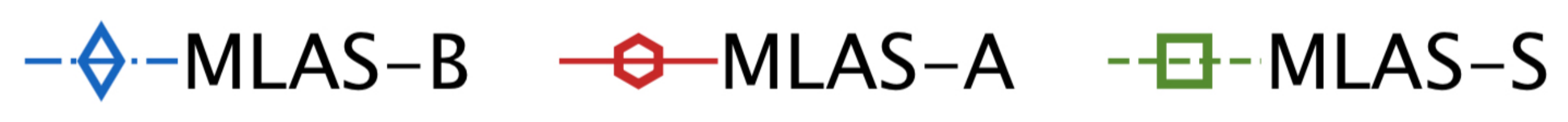}
  \vspace{-1mm}
  
      \begin{subfigure}[t]{0.24\linewidth}
        \centering
        \includegraphics[page=1, width=\linewidth]{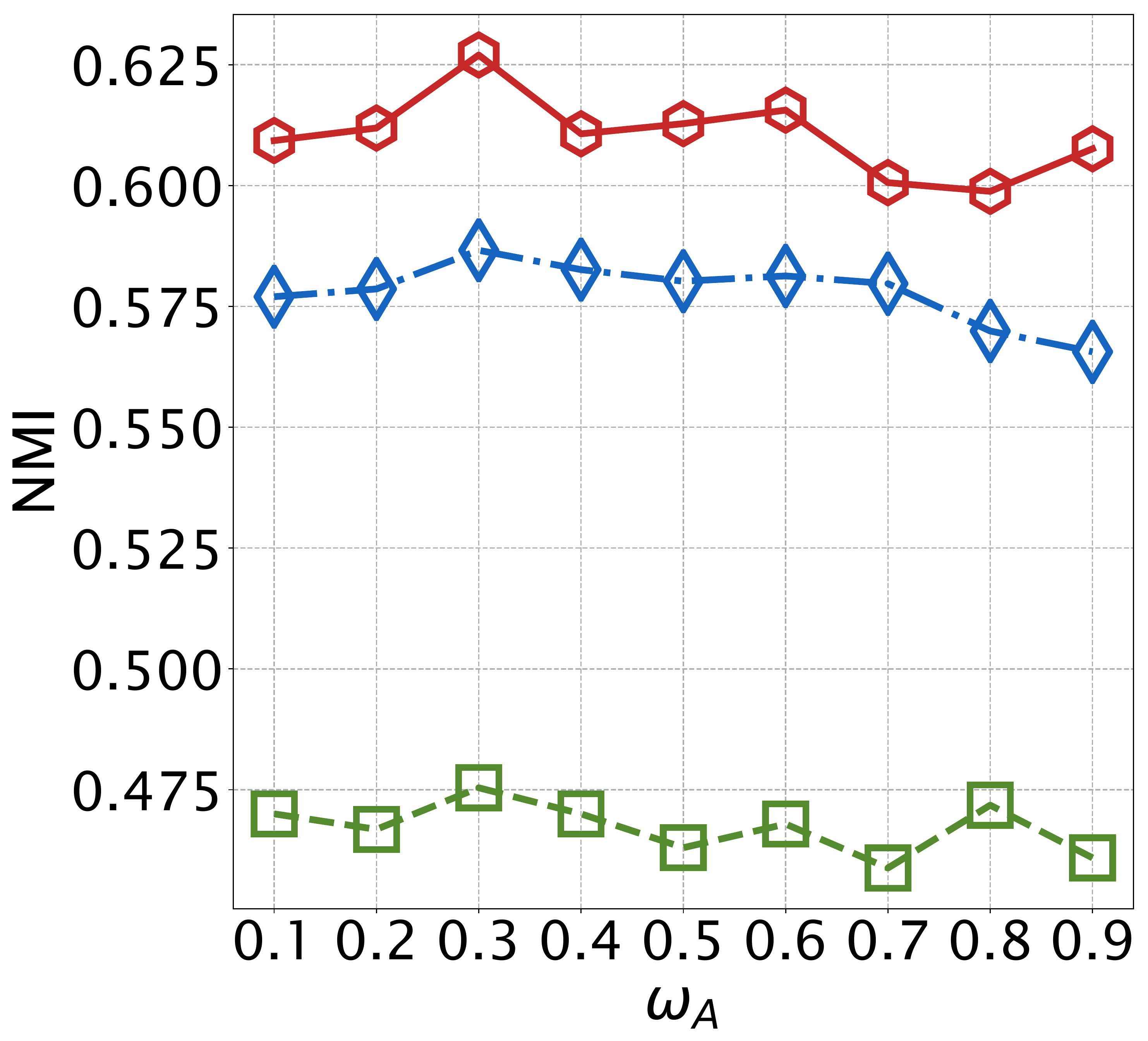}
        \label{fig-exp-w1}
        \vspace{-5mm}
        \caption{AMS-A Dataset}
      \end{subfigure}
      \hfill
      \begin{subfigure}[t]{0.24\linewidth}
        \centering
        \includegraphics[page=2, width=\linewidth]{nmi_w}
        \label{fig-exp-w2}
        \vspace{-5mm}
        \caption{AMS-B Dataset}
      \end{subfigure}
      \hfill
      \begin{subfigure}[t]{0.24\linewidth}
        \centering
        \includegraphics[page=3, width=\linewidth]{nmi_w}
        \label{fig-exp-w3}
        \vspace{-5mm}
        \caption{Wiki-A Dataset}
      \end{subfigure}
      \hfill
      \begin{subfigure}[t]{\figwidth\linewidth}
        \centering
        \includegraphics[page=4, width=\linewidth]{nmi_w}
        \label{fig-exp-w4}
        \vspace{-5mm}
        \caption{Wiki-B Dataset}
      \end{subfigure}
      \vspace{-3mm}
     \caption{The effect of pre-training parameters in \sysnamenospace. The pre-training parameter $\omega_A$ decides the relative importance of attributes in the model. We observe that \sysnamenospace-\texttt{A} is capable of achieving the best performance on AMS-A and AMS-B datasets while \sysnamenospace-\texttt{S} has the best performance on Wiki-A and Wiki-B datasets. }
    \label{fig-exp-wa}
    \vspace{-3mm}
  \end{figure*} 
\subsubsection{Performance in Clustering Tasks.} The primary parameter in HDBSCAN is the minimum cluster size~\cite{campello2015hierarchical}, denoting the smallest set of instances to be considered as a group. 
Intuitively, while the minimum cluster size increases, each cluster may include instances that do not belong to it and the performance decreases. 
Figure~\ref{fig-exp-clustering} presents the results with the output dimension is 10 and $\omega_A = 0.5$. 

Compared to the best baseline method \texttt{ASF}, \sysnamenospace-\texttt{A} achieves up to 18.3\% and 25.4\% increase of performance on AMS-A and AMS-B datasets, respectively. On Wiki-A and Wiki-B datasets, \sysnamenospace-\texttt{S} is capable of achieving up to 26.3\% and 24.8\% performance improvement compared to \texttt{ASF}, respectively. 
We further confirm that \sysname network is capable of exploiting the \textit{attribute-sequence dependencies} to improve the performance of the clustering algorithm with various parameter settings. 
\subsubsection{Output Dimensions.} We evaluate \sysname under a wide range of output dimension choices. The number of output dimensions relates to a variety of impacts, such as the usability of feature representations in downstream data mining tasks. In this set of experiments, we fix the minimum cluster size at 50, $\omega_A=0.5$ and vary output dimensions from 10 to 100. From Figure~\ref{fig-exp-dimension} we conclude that our proposed approaches outperform the baseline methods with various output dimensions. 

In particular, compared to the baseline method with the best performance, namely \texttt{ASF}, \sysnamenospace-\texttt{A} achieves 20.7\% improvement on average on AMS-A dataset and 19.4\% improvement on average on the AMS-B dataset. When evaluated using Wiki-A and Wiki-B datasets, \sysnamenospace-\texttt{S} outperforms \texttt{ASF} by 20.8\% and 10.6\% on average. 

\begin{figure*}[t]
\centering
    \begin{subfigure}{\linewidth}
    \centering
      \includegraphics[width=0.75\linewidth]{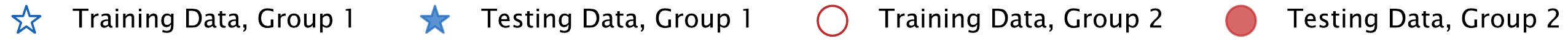}
    \end{subfigure}
    \begin{subfigure}[t]{0.15\linewidth}
      \centering
      \includegraphics[page=1, width=\linewidth]{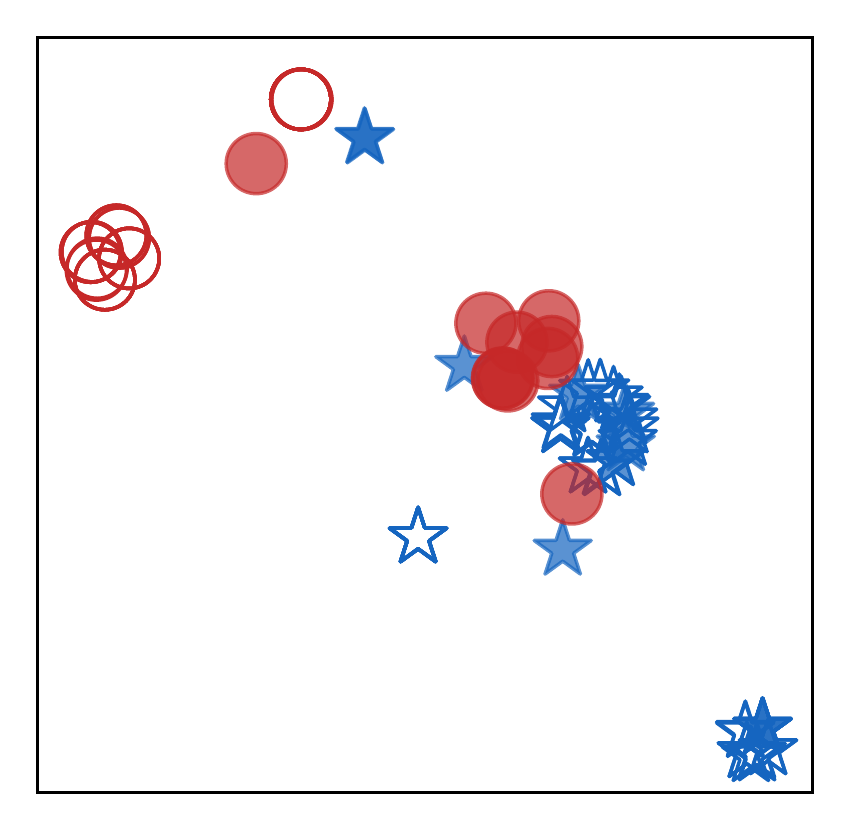}
      \captionsetup{format=centercaption}
      \caption{\texttt{ATT} \\ without Feedback}
      \label{fig-exp-tsne-attn}
      \end{subfigure}
    \hfill
    \begin{subfigure}[t]{0.15\linewidth}
      \centering
      \includegraphics[page=3, width=\linewidth]{tsne_plot}
        \captionsetup{format=centercaption}
      \caption{\texttt{SEQ} \\ without Feedback}
      \label{fig-exp-tsne-seqn}
      \end{subfigure}
    \hfill
    \begin{subfigure}[t]{0.15\linewidth}
      \centering
      \includegraphics[page=11, width=\linewidth]{tsne_plot}
        \captionsetup{format=centercaption}
      \caption{\texttt{ASF} \\ without Feedback}
      \label{fig-exp-tsne-asfn}
      \end{subfigure}
    \hfill
    \begin{subfigure}[t]{0.15\linewidth}
      \centering
      \includegraphics[page=9, width=\linewidth]{tsne_plot}
        \captionsetup{format=centercaption}
      \caption{\sysnamenospace\texttt{-B} \\ without Feedback}
      \label{fig-exp-tsne-fcn}
      \end{subfigure}
    \hfill
    \begin{subfigure}[t]{0.15\linewidth}
      \centering
      \includegraphics[page=5, width=\linewidth]{tsne_plot}
        \captionsetup{format=centercaption}
      \caption{\sysnamenospace\texttt{-A} \\ without Feedback}
      \label{fig-exp-tsne-fan}
      \end{subfigure}
    \hfill
    \begin{subfigure}[t]{0.15\linewidth}
      \centering
      \includegraphics[page=7, width=\linewidth]{tsne_plot}
        \captionsetup{format=centercaption}
      \caption{\sysnamenospace\texttt{-S} \\ without Feedback}
      \label{fig-exp-tsne-fsn}
      \end{subfigure}
    \hspace{-2mm}
    \begin{subfigure}[t]{0.15\linewidth}
      \centering
      \reflectbox{\includegraphics[page=2, width=\linewidth]{tsne_plot}}
      \captionsetup{format=centercaption}
      \caption{\texttt{ATT} \\ with Feedback}
      \label{fig-exp-tsne-attf}
      \end{subfigure}
    \hfill
    \begin{subfigure}[t]{0.15\linewidth}
      \centering
      \includegraphics[page=4, width=\linewidth]{tsne_plot}
      \captionsetup{format=centercaption}
      \caption{\texttt{SEQ} \\ with Feedback}
      \label{fig-exp-tsne-seqf}
      \end{subfigure}
    \hfill
    \begin{subfigure}[t]{0.15\linewidth}
      \centering
      \includegraphics[page=12, width=\linewidth]{tsne_plot}
      \captionsetup{format=centercaption}
      \caption{\texttt{ASF} \\ with Feedback}
      \label{fig-exp-tsne-asff}
      \end{subfigure}
    \hfill
    \begin{subfigure}[t]{0.15\linewidth}
      \centering
      \includegraphics[page=10, width=\linewidth]{tsne_plot}
      \captionsetup{format=centercaption}
      \caption{\sysnamenospace\texttt{-B} \\ with Feedback}
      \label{fig-exp-tsne-fcf}
    \end{subfigure}
    \hfill
    \begin{subfigure}[t]{0.15\linewidth}
      \centering
      \reflectbox{\includegraphics[page=6, width=\linewidth]{tsne_plot}}
      \captionsetup{format=centercaption}
      \caption{\sysnamenospace\texttt{-A} \\ with Feedback}
      \label{fig-exp-tsne-faf}
      \end{subfigure}
    \hfill
    \begin{subfigure}[t]{0.15\linewidth}
      \centering
      \includegraphics[page=8, width=\linewidth]{tsne_plot}
      \captionsetup{format=centercaption}
      \caption{\sysnamenospace\texttt{-S} \\ with Feedback}
      \label{fig-exp-tsne-fsf}
      \end{subfigure}
    \hfill
  \caption{Plots of the feature representations. The \sysname is capable of exploiting the feedback and separating the instances from two different groups while keeping the instances from the same group together. }%
  \label{fig-exp-tsne}
  \vspace{-5mm}
\end{figure*} 
\subsubsection{Pre-training Parameters} We evaluate \sysname under different pre-training parameters in this set of experiments. 
\texttt{ATT} and \texttt{SEQ} are not included in this set of experiments since they only utilize one data type. Output dimension is set to 5. Minimum cluster size is set at 50. 
Figure~\ref{fig-exp-wa} presents the results under different pre-training parameters. This confirms that our proposed \sysname method is not sensitive to different pre-training parameters. 

We notice the performance differences among the three \sysname architectures in the above experiments, where \sysnamenospace-\texttt{A} has the best performance on AMS-A and AMS-B datasets, and \sysnamenospace-\texttt{S} has the best performance on Wiki-A and Wiki-B datasets. We conclude this difference may relate to the datasets. 

\subsection{Case Studies} 
In Figure~\ref{fig-exp-tsne}, we apply t-SNE~\cite{maaten2008visualizing} to the feature representations generated by all compared methods. The set of feature representations without feedback is generated after the pre-training phase and before the distance metric learning process. 

Our goal is to demonstrate the differences in the feature space of each method. 
We randomly select data points from both training and testing sets with a ground truth of two groups. We have the following findings: 

1. The methods using either attribute data (\texttt{ATT}) or  sequence data (\texttt{SEQ}) \textit{only} cannot use the attributed sequence feedback.

2. The method using \textit{both} attributes and sequences \textit{separately} (\texttt{ASF}) is capable of better separating the two groups than the methods using single data type (\texttt{ATT} and \texttt{SEQ}). 

3. Our methods using attributed sequence feedback as a unity to train unified models (\sysnamenospace-\texttt{B/A/S}) are capable of separating the two groups the furthest, and thus achieve the best results. 

These observations confirm that all three designs of \sysname can effectively learn the distance metric and result in better separation of two groups of data points. 

\section{Related Work}
\label{section-related}
\subsection{Distance Metric Learning}
Distance metric learning, with the goal of learning a distance metric from pairs of similar and dissimilar examples, has been extensively studied ~\cite{xing2003distance,yeung2007kernel,davis2007information,wang2011integrating,mignon2012pcca, koestinger2012large,cvpr-face-verify,mueller2016siamese,neculoiu2016learning}. The common objective of these tasks is to learn a distance metric so that the distance between similar pairs is reduced while the distance between dissimilar pairs is enlarged. Distance metric learning has been used  to improve the performance of mining tasks, such as clustering~\cite{xing2003distance,yeung2007kernel,davis2007information}. Many application domain require distance metric learning, including patient similarity in health informatics~\cite{wang2011integrating}, face verification in computer vision~\cite{mignon2012pcca,koestinger2012large, cvpr-face-verify}, image recognition~\cite{koch2015siamese} and sentence semantic similarity analysis~\cite{mueller2016siamese, neculoiu2016learning}. 
 Recent works on distance metric learning using deep learning techniques have been using using siamese neural networks with two identical base networks in various supervised tasks~\cite{koch2015siamese,mueller2016siamese,cvpr-face-verify,zhuang2018one}. 
However, each of these works~\cite{koch2015siamese,mueller2016siamese,cvpr-face-verify} focuses on a domain-specific problem with a homogeneous data type. Thus, the dependency challenge remain unexplored in these works. 
  
\subsection{Deep Learning} Deep learning has received significant interests in recent years. Deep learning models are capable of feature learning with varying granularities. Various deep learning models and optimization techniques have been proposed in a wide range of applications from image recognition~\cite{karpathy2015deep,xu2015show} to sequence learning~\cite{cho-EtAl:2014, sutskever2014sequence, xu2017decoupling, neculoiu2016learning,zhuang2019amas}. Many applications involve the learning of a single data type~\cite{cho-EtAl:2014, sutskever2014sequence, xu2017decoupling,neculoiu2016learning}, while some applications involve more than one data type~\cite{karpathy2015deep,xu2015show}. Several works~\cite{cvpr-face-verify,mueller2016siamese,neculoiu2016learning} focus on deep metric learning using deep learning techniques. However, none of these works address the deep metric learning of more than one type of data nor the dependencies between different types of data. 
\section{Conclusion and Future Work}
\label{section-conclusion}
In this work, we focus on the novel problem of distance metric learning on attributed sequences. We first identify and formally define this prevalent data type of attributed sequences and the problem. We then propose one \sysname with three solution variations to this problem using neural network models. The proposed \sysname network effectively learns the nonlinear distance metric from both attribute and sequence data, as well as the attribute-sequence dependencies. In our experiments on real-world datasets, we demonstrate the effectiveness of our \sysname network over other state-of-the-art methods in both performance evaluations and case studies. 

The prevalence of attributed sequence data and the broad spectrum of real-world applications using attributed sequences motivate us to keep exploring this new direction of research. 
Given the performance boost using attributed sequences and different variations of neural network models, future research could focus on exploring different design choices of the \mnetns. It would also be interesting to explore the design of alternative neural network models as the components within \sysnamenospace, such as \anet and \snetns. Another research topic would be the theoretical analysis of the performance differences among the three \sysname network architectures.

  \balance
  \bibliographystyle{IEEEtran}%
  \bibliography{referencesfull}
  
  \end{document}